\documentclass[letterpaper]{article} 
\usepackage[draft]{aaai25}
\usepackage{times}
\usepackage{helvet} 
\usepackage{courier}
\usepackage[citestyle=numeric,
    style=numeric,
    firstinits=false,
    maxnames=100,
    sorting=none]{biblatex}
\usepackage{graphicx} 
\usepackage{caption}
\usepackage{algorithm}
\usepackage{algorithmic}

\usepackage{newfloat}
\usepackage{listings}
\DeclareCaptionStyle{ruled}{labelfont=normalfont,labelsep=colon,strut=off} % DO NOT CHANGE THIS
\floatstyle{ruled}
\newfloat{listing}{tb}{lst}{}
\floatname{listing}{Listing}

\title{Efficient Federated Finetuning of Tiny Transformers with Resource-Constrained Devices}
\author{
    Kilian Pfeiffer\textsuperscript{\rm 1},
    Mohamed Aboelenien Ahmed\textsuperscript{\rm 1},
    Ramin Khalili\textsuperscript{\rm 2},
    Jörg Henkel\textsuperscript{\rm 1}
}
\affiliations{
    \textsuperscript{\rm 1}Karlsruhe Institute of Technology,
    Karlsruhe, Germany\\
    \textsuperscript{\rm 2}Huawei Research Center,
    Munich, Germany\\
    kilian.pfeiffer@kit.edu, mohamed.ahmed3@kit.edu, ramin.khalili@huawei.com, henkel@kit.edu\\
}

\usepackage{tikz}
\usepackage{pgfplots}
\usepackage{amsmath}
\usetikzlibrary{plotmarks}
\usepgfplotslibrary{fillbetween}
\usetikzlibrary{arrows, decorations.markings, decorations}
\usetikzlibrary{arrows,arrows.meta,patterns,positioning,shapes}
\usetikzlibrary{plotmarks}
\usepgfplotslibrary{fillbetween}
\usetikzlibrary{arrows, decorations.markings, decorations}
\usepackage{mathtools}
\usepackage{acro}
\usepackage{booktabs}
\usepackage{pgfplots}
\usepgfplotslibrary{groupplots,dateplot}
\usepackage[all]{nowidow}
\usepackage{soul}
\usepackage{tabularx,multirow}
\usepackage{xspace}
\usepackage{tikz}
\usepackage{bm}
\usepackage[flushleft]{threeparttable}
\usetikzlibrary{arrows,arrows.meta,patterns,positioning,shapes}
\usepackage{soul}
\usepackage{enumitem}
\usepackage{makecell}
\usepackage{siunitx}
\usepackage{multirow}
\usepackage{adjustbox}
\usepackage{csvsimple}

\pgfplotsset{
    compat=1.15,
}

\definecolor{tabblue}{HTML}{5075b2}
\definecolor{tabred}{HTML}{bf575a}
\definecolor{tabgreen}{HTML}{71b582}
\definecolor{tabyellow}{HTML}{cdba77}
\DeclareSIUnit{\pp}{\textup{p.p.}}

\usepackage{amssymb}
\usepackage{hyperref}
\usepackage{cleveref}
\usepgfplotslibrary{groupplots,units}
\usepackage{acro}
\DeclareAcronym{FL}{short=FL,long=federated learning,short-indefinite=an}
\DeclareAcronym{IOT}{short=IoT,long=internet of things,short-indefinite=an}
\DeclareAcronym{NN}{short=NN,long=neural network,short-indefinite=an}
\DeclareAcronym{CNN}{short=CNN, long=convolutional neural network}
\DeclareAcronym{FEDAVG}{short=FedAvg, long=Federated Averaging}
\DeclareAcronym{IID}{short=iid, long=independent and identically distributed}
\DeclareAcronym{ML}{short=ML, long=machine learning}
\DeclareAcronym{KD}{short=KD, long=knowledge distillation}
\DeclareAcronym{SGD}{short=SGD, long=stochastic gradient descent}
\DeclareAcronym{NAS}{short=NAS, long=neural architecture search}
\DeclareAcronym{FD}{short=FD, long=Federated Dropout}
\DeclareAcronym{FLOP}{short=FLOP, long=Floating Point Operation}
\DeclareAcronym{LLM}{short=LLM, long=Large Language Model}
\DeclareAcronym{SVD}{short=SVD, long=singular value decomposition}
\DeclareAcronym{VIT}{short=ViT, long=Vision Transformer}
\newcommand{\etal}{\emph{et al.}\xspace}

\DeclareAcronym{OURS}{short=CAFF, long=Contraint-Aware Federated Finetuning}
\begin{document}

\maketitle

\begin{abstract}
In recent years, \acp{LLM} through Transformer structures have dominated many machine learning tasks, especially text processing. However, these models require massive amounts of data for training and induce high resource requirements, particularly in terms of the large number of \acp{FLOP} and the high amounts of memory needed. To fine-tune such a model in a parameter-efficient way, techniques like Adapter or LoRA have been developed. However, we observe that the application of LoRA, when used in \ac{FL}, while still being parameter-efficient, is memory and \ac{FLOP} inefficient. Based on that observation, we develop a novel layer finetuning scheme that allows devices in cross-device \ac{FL} to make use of pretrained \acp{NN} while adhering to given resource constraints. We show that our presented scheme outperforms the current state of the art when dealing with homogeneous or heterogeneous computation and memory constraints and is on par with LoRA regarding limited communication, thereby achieving significantly higher accuracies in \ac{FL} training.
\end{abstract}

\acresetall

\section{Introduction}
In recent years, \acp{LLM} have dominated various machine learning tasks, particularly next token prediction and text classification, while \acp{VIT} have closed the gap with \acp{CNN} in vision tasks. However, these models require massive amounts of data (e.g., huge quantities of text for \acp{LLM}) and are very power-hungry for training, as they have billions of parameters to adjust~\cite{li2020train}.
As these \acp{LLM}, like GPT2~\cite{radford2019language} or LLaMA~\cite{touvron2023llama}, have billions of parameters and are trained on large quantities of text, they generalize well to many downstream tasks in the text domain.
Similarly, in vision, multimodal Transformers can generalize well to detecting objects in an image \cite{hu2021unit, wang2022git}.

\begin{figure*}[h]
    \centering
    \includegraphics[page=1]{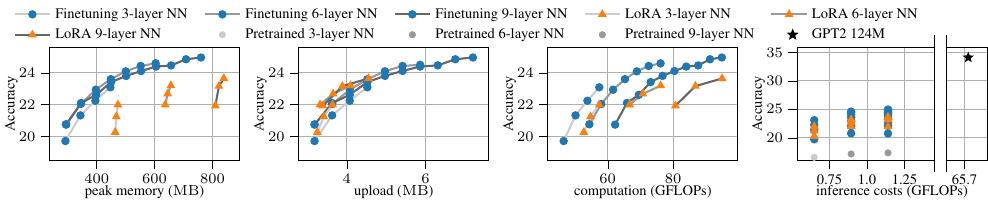}
    \caption{Comparison of downstream performance and resource requirements for next-token prediction on Shakespeare~\cite{caldas1812leaf} using layer finetuning (with layers frozen from first to last, where each dot represents a specific number of layers being frozen) and LoRA with tiny Transformers pretrained on OpenWebText, having 3, 6, and 9 layers. We observe that while LoRA (with ranks 24, 12, 3) can achieve gains in communication efficiency, it requires significantly more peak memory and \acp{FLOP} to reach the same accuracy. Hyperparameters and details are provided in \cref{subsec:hyperparameters}. GPT2 is included to highlight inference costs (accuracy is based on GPT2's tokenizer and test data may be part of GPT2's training set).}
    \label{fig:motivation}
\end{figure*}

In many downstream applications, e.g., in next-word prediction or object classification on smartphones or \ac{IOT} devices, deploying such large models imposes high resource requirements for inference, potentially causing high latency and high energy consumption.
Although, through generalization, they can achieve the desired accuracy, for many tasks, they are not necessarily required, as tiny specialized models would suffice. In particular, we observe that tiny models that require~$\sim 60\times$ fewer resources (\acp{FLOP}) for inference compared to a lightweight GPT-2 124M model can perform similarly\footnote{Parts of the test set may be in GPT's training set, hence increasing accuracy.}
in next-token prediction tasks (refer to \cref{fig:motivation}).
However, these specialized models require a sufficient amount of problem-specific data to serve as a replacement. In many cases, this problem-specific data resides on \textit{resource-constrained} edge devices (such as smartphones or \ac{IOT} devices), is privacy-sensitive, and cannot be centrally stored and processed. While in centralized training, several hundred-watt server GPUs are available, edge devices are very limited in their resources and can only spend a few watts on training. Additionally, such devices can have \emph{heterogeneous} capabilities~\cite{pfeiffer2023federated}.

To make use of the devices' data, training must be performed on the devices themselves. In recent years, \ac{FL} has emerged as a privacy-sensitive alternative to centralized training and has shown success in many domains such as smartphone applications, healthcare, and robotics~\cite{chen2020fedhealth, yuan2020federated, ciftler2020federated, posner2021federated}.
In this work, we study how tiny pretrained Transformer models can be adapted to downstream tasks using \ac{FL} with heterogeneous resource-constrained devices.

To adapt large models to downstream tasks, recently, popular techniques like Adapter~\cite{houlsby2019adapter} and LoRA~\cite{hu2021lora} have been introduced, mainly allowing the adaptation of such models in a parameter-efficient way.
However, we observe that while being parameter-efficient, techniques like LoRA still require massive amounts of memory for training in the case of tiny models.
The reason for this is that LoRA mainly reduces the memory overhead of gradients and optimizer states but does not lower the activation memory.
For large models, this typically suffices as the weights and gradients account for most of the required training memory.
We observe that the memory footprint of tiny language models and \acp{VIT} is mainly dominated by the activation memory (\cref{appdx:technical_details}).

We compare LoRA against finetuning individual layers of a set of tiny \acp{NN} (Transformers with $3$–$12$ layers, $3$ heads, and an embedding size of $92$) that were pretrained on OpenWebText~\cite{Gokaslan2019OpenWeb} and trained in a federated manner to perform next-token prediction on Shakespeare~\cite{caldas1812leaf} from the Leaf benchmark. From~\cref{fig:motivation}, we can draw the following conclusion: For tiny language models, LoRA can achieve similar communication savings compared to layer finetuning.
With LoRA, only the low-rank adapters need to be uploaded, while with layer finetuning, only the trained layers need to be uploaded.
However, besides memory, we observe that compared to layer finetuning, LoRA requires significantly more computation (as backpropagation is needed from the last to the first layer).
Motivated by that observation, we re-evaluate existing \ac{FL} strategies that address resource constraints and heterogeneous devices in cross-device \ac{FL}.

Different from the state of the art, we assume that the \ac{FL} models are already pretrained. Furthermore, we assume that we have a selection of differently sized \acp{NN} to choose from.
We observe that when using pretrained \acp{NN}, layer finetuning outperforms LoRA-derived techniques~\cite{cho2023heterogeneous} as well as existing state-of-the-art methods~\cite{yao2021fedhm, alam2022fedrolex, diao2020heterofl, horvath2021fjord, kim2023depthfl}. Based on our observations, we propose a strategy that, through \ac{NN} selection and layer finetuning, allows reaching the highest accuracy while adhering to a given device constraint.
In summary, we make the following novel contributions:

\begin{itemize}
    \item We are the first to study resource-constrained cross-device \ac{FL} with the availability of pretrained models. We rigorously evaluate existing \ac{FL} strategies and observe that layer finetuning outperforms vanilla LoRA~\cite{hu2021lora} as well as recent \ac{FL} adaptations for heterogeneity~\cite{cho2023heterogeneous}. Furthermore, with pretrained tiny \ac{NN} models, layer finetuning also surpasses existing \ac{FL} techniques~\cite{diao2020heterofl, alam2022fedrolex, horvath2021fjord, yao2021fedhm, kim2023depthfl}. Specifically, we study \ac{FL} downstream tasks such as Shakespeare, CIFAR10, and CIFAR100.
    \item Existing works assume a fixed \ac{NN} that must be trained while adhering to device constraints. In this work, we assume a set of pretrained \ac{NN} architectures is available for selection. Based on our observations, we propose \acf{OURS}, an architecture selection technique that achieves the highest accuracy given a device constraint.
    \item We evaluate the performance of state-of-the-art techniques and \ac{OURS} in heterogeneous settings. Additionally, our technique promotes greater fairness for weaker devices.
\end{itemize}

\section{Related Work}
\textbf{Subset-based training:} Resource constraints, and especially heterogeneous devices in \ac{FL}, are tackled in a variety of works. A large branch of works applies \textit{submodel} training~\cite{caldas2018expanding, horvath2021fjord, diao2020heterofl, alam2022fedrolex}. Caldas~\etal\cite{caldas1812leaf} were among the first to propose such a scheme. Specifically, they randomly drop filters in \acp{CNN} to create a submodel with lower resource intensity. Both schemes in HeteroFL~\cite{diao2020heterofl} and FjORD~\cite{horvath2021fjord} hierarchically drop filters, such that devices with specific constraints always receive the same parameters. Additionally, in FjORD, each device alternates between subsets that are within its capabilities. However, both require that the most capable device is able to train the full \ac{NN} model. FedRolex~\cite{alam2022fedrolex} addresses this issue by applying a rolling-window scheme over the parameter set, such that eventually all parameters receive training. Lastly, DepthFL~\cite{kim2023depthfl} trains subsets by splitting the \ac{NN} depthwise, using early exits for constrained devices in combination with self-distillation. Beyond training of subsets, freezing and quantization have been considered to address the heterogeneity challenges.

\textbf{Freezing in \ac{FL}:} In CoCoFL~\cite{pfeiffer2023cocofl}, freezing and quantization are used to lower the resources in heterogeneous training. However, it is assumed that the \ac{NN} model is trained from scratch, i.e., there has to be a set of devices capable of training the first layers. Hence, the selection technique aims to propagate the gradients as far back as possible within the \ac{NN} structure. Freezing has also been used to progressively increase the size of the \ac{NN} and thereby save resources. Wang~\etal propose ProgFed~\cite{wang2022progfed}, where the \ac{NN} is progressively increased by stacking layers on top of each other. Each time a new layer is stacked on the previously trained \ac{NN}, the \ac{NN}'s head is removed and replaced. The authors show that this enables a significant reduction in required data upload and computations. However, eventually devices have to train the fully stacked \ac{NN} as no freezing is applied, thereby requiring memory capabilities for fully training the \ac{NN} model. This constraint is loosened by Successive Layer Training~\cite{pfeiffer2023aggregating}, where a model is progressively built up, but more and more early layers get frozen; thereby, the technique allows to obey to a given memory constraint. However, both techniques assume training a model in a federated fashion from scratch and do not consider pretrained models. Freezing in \ac{FL} has also been used to tackle other problems, like personalization~\cite{setayesh2023perfedmask}. Lastly, freezing schemes have been proposed to tackle communication efficiency~\cite{malan2023automatic}.

\textbf{Low-rank factorization:} Low-rank factorization has been considered in tackling heterogeneity~\cite{yao2021fedhm, mei2022resource} in \ac{FL}. Similar to LoRA~\cite{hu2021lora}, low-rank adapters are used, i.e., a layer that maps from the input space to a lower-rank space and a second layer that maps from the lower-rank space to the output space. Differently from LoRA, these techniques replace the \ac{NN}'s main parameters using a lower-rank approximation gathered through singular value decomposition.

\textbf{LoRA in federated training of \acp{LLM}:} With the increasing popularity of \acp{LLM}, LoRA has been proposed to train such models in federated \textit{cross-silo} scenarios~\cite{babakniya2023slora, yu2020heterogeneous}. However, the main focus of existing works is fine-tuning large \ac{NN} models (e.g., GPT2, LLaMA) in a communication-efficient way, still imposing large memory and computation requirements on devices. In \cite{yu2020heterogeneous}, communication heterogeneity is tackled, where LoRA layers with different ranks are distributed to heterogeneous devices. Aggregation of the LoRA layers is performed similarly to HeteroFL.

\textit{In summary, apart from recent works in fine-tuning \acp{LLM}, none of the existing works consider the availability of a set of pretrained \ac{NN} models. Existing LoRA techniques~\cite{babakniya2023slora, yu2020heterogeneous} focus on cross-silo communication and neglect memory and computation constraints in case of tiny Transformers.}

\section{Methodology}

\textbf{Problem statement:} We consider \iac{FL} (synchronous cross-device) problem, where a single \textit{server} and many \textit{devices}~$c \in \mathcal{C}$ exist. Training is done on the devices for several rounds~$R$, where the server is only responsible for aggregation. Furthermore, we require that a set of pretrained tiny Transformer architectures~$\mathcal{F}$ exists, where each architecture~$F_l \in \mathcal{F}$ has a specific number of stacked layers~$l$ (specifically, a layer constitutes a multi-head attention module, as well as feedforward blocks). We assume all \acp{NN} in this set satisfy inference latency requirements. A fixed number of devices~$|\mathcal{C}^{(r)}| \ll |\mathcal{C}|$ participate every round~$r \leq R$.
Further, we assume that devices are subject to constraints. Specifically, we assume that a device can only train with a limited amount of \textit{memory}, can be restricted in its \textit{upload} capabilities\footnote{We assume that only limited bandwidth is available for a device to upload its weights within a reasonable time within a round.}, and can only perform a limited number of \textit{computations} (\acp{FLOP}) per round. Without loss of generality, we assume that these constraints are known to the server and are static over time. We label~$M_c$ as the peak memory available for training on device~$c$. Similarly, we label~$U_c$ and~$O_c$ as the amount of upload a device can do and the computation that can be performed, respectively. Consequently, a device~$c$ can only participate in the training if the selected \ac{NN}-\textit{configuration} satisfies the given constraints.

We aim to maximize the accuracy of the \ac{FL} system, given a set of devices $\mathcal{C}$, with a set of (heterogeneous) constraints $\mathcal{U}=\{U_c : \forall c \in \mathcal{C}\}$, $\mathcal{M}=\{M_c : \forall c \in \mathcal{C}\}$, and $\mathcal{O}=\{O_c : \forall c \in \mathcal{C}\}$ within a limited number of rounds $R$.

\begin{algorithm}[tb]
\caption{\acs{OURS}}
\label{alg:algorithm}
\textbf{Input}: Total rounds~$R$, devices~$\mathcal C$, number of devices per round~$|\mathcal{C}^{(r)}|$, device constraints $\mathcal{M}$, $\mathcal{U}$, $\mathcal{O}$, set of pretrained \acp{NN}\\
\begin{algorithmic}[1] 
\STATE $\mathcal{F}_{\text{feasible}} \gets \{ F_l \in \mathcal{F} \ | \ (\exists t) [t \in [1,\ldots,l] \land \forall c \in \mathcal C [M_{F_l^t} \leq M_c  \land U_{F_l^t} \leq U_c  \land O_{F_l^t} \leq O_c ]]\}$
\STATE $\mathcal{W}^{(1)} \gets \mathcal{W}_{\text{pretrained}} \gets F_l$ \texttt{// get pretrained}
\FORALL{\normalfont{round} $r=1,2,\ldots, R$}
\STATE $\mathcal{C}^{(r)}$ $\gets$ select $|\mathcal{C}^{(r)}|$ devices randomly out of $\mathcal C$
\FOR{\normalfont{device} $c \in \mathcal C^{(r)}$ in parallel}
\STATE $\mathcal W^{(r)}$ receive from server
\STATE $\mathcal W^{(r,c)} \gets$ \textbf{LocalTraining}$(\mathcal W^{(r)})$
\STATE upload $\mathcal W^{(r,c)}$
\ENDFOR
\FOR{\normalfont{layer} $i \in [1,\ldots,l]$}
\STATE $\tilde{\mathcal W}_i \gets \bigcup_{c \in \mathcal C} \{W_i^{(r,c)} \ | \ \exists W_i^{(r,c)} \in \mathcal{W}^{(r,c)} \}$
\STATE $W_i^{(r+1)} \gets \frac{1}{|\tilde{\mathcal W}_i|} \big(\sum_{W_j \in \tilde{\mathcal W}_i} W_j \big) + (1-\frac{1}{|\tilde{\mathcal W}_i|})\cdot W_i^{(r)}\quad$\texttt{// weighted averaging}
\ENDFOR
\ENDFOR
\STATE \phantom{a}
\STATE \textbf{LocalTraining} $(\mathcal W)$
\STATE train $W_i$ for $i\in [l-t_c+1,\ldots,l]$ based on $M_c, U_c, C_c$
\STATE \textbf{Return:} $\mathcal W \gets \{W_i \ | \ \forall i \in [l-t_c+1,\ldots,l]\}$
\end{algorithmic}
\end{algorithm}

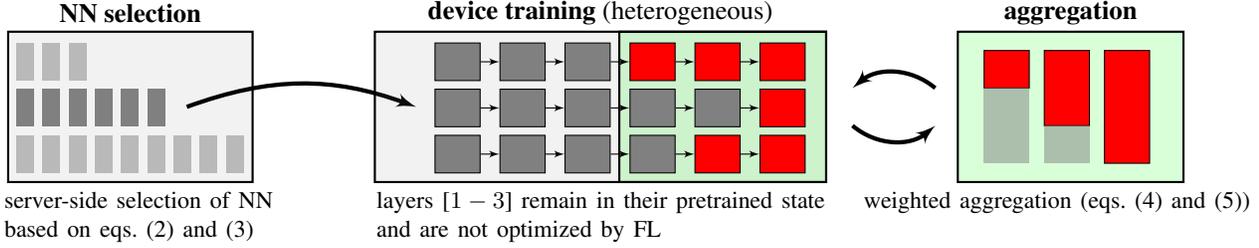
\begin{figure*}[h]
    \centering
    \begin{tikzpicture}[node distance=1cm,auto,>=latex']	
	\definecolor{tempcolor}{HTML}{009682}
	
	\begin{scope}[node distance=1.25cm,auto,>=latex', shift={(1.5,0)}]
		\tikzstyle{block}=[draw, fill=blue!20, minimum width=8mm,%
		minimum height=1cm]

		\node[yshift=1.25cm] (){\textbf{NN selection}};
		\node[text width=3.6cm, anchor=north west, yshift=-1.0cm, xshift=-1.8cm] (){\small server-side selection of \ac{NN} based on~\cref{eq:feasible,eq:min2}};

		\node[draw, thick, minimum width=3.25cm, minimum height=2cm, fill=gray, fill opacity=0.1](){};

		\node[fill opacity=0.5, minimum width=0.2cm, minimum height=0.5cm, fill=gray] at (-1.4,0.6) (a1) {};
		\node[fill opacity=0.5, minimum height=0.5cm, fill=gray, right=0.1cm of a1] (a2) {};
		\node[fill opacity=0.5, minimum height=0.5cm, fill=gray, right=0.1cm of a2] (a3) {};

		\node[minimum width=0.2cm, minimum height=0.5cm, fill=gray, below=0.1cm of a1] (b1) {};
		\node[minimum width=0.2cm, minimum height=0.5cm, fill=gray, right=0.1cm of b1] (b2) {};
		\node[minimum width=0.2cm, minimum height=0.5cm, fill=gray, right=0.1cm of b2] (b3) {};
		\node[minimum width=0.2cm, minimum height=0.5cm, fill=gray, right=0.1cm of b3] (b4) {};
		\node[minimum width=0.2cm, minimum height=0.5cm, fill=gray, right=0.1cm of b4] (b5) {};
		\node[minimum width=0.2cm, minimum height=0.5cm, fill=gray, right=0.1cm of b5] (b6) {};

		\node[fill opacity=0.5, minimum width=0.2cm, minimum height=0.5cm, fill=gray, below=0.1cm of b1] (c1) {};
		\node[fill opacity=0.5, minimum width=0.2cm, minimum height=0.5cm, fill=gray, right=0.1cm of c1] (c2) {};
		\node[fill opacity=0.5, minimum width=0.2cm, minimum height=0.5cm, fill=gray, right=0.1cm of c2] (c3) {};
		\node[fill opacity=0.5, minimum width=0.2cm, minimum height=0.5cm, fill=gray, right=0.1cm of c3] (c4) {};
		\node[fill opacity=0.5, minimum width=0.2cm, minimum height=0.5cm, fill=gray, right=0.1cm of c4] (c5) {};
		\node[fill opacity=0.5, minimum width=0.2cm, minimum height=0.5cm, fill=gray, right=0.1cm of c5] (c6) {};
		\node[fill opacity=0.5, minimum width=0.2cm, minimum height=0.5cm, fill=gray, right=0.1cm of c6] (c7) {};
		\node[fill opacity=0.5, minimum width=0.2cm, minimum height=0.5cm, fill=gray, right=0.1cm of c7] (c8) {};
		\node[fill opacity=0.5, minimum width=0.2cm, minimum height=0.5cm, fill=gray, right=0.1cm of c8] (c9) {};

		\draw[->, ultra thick] (0.75,0) to[out=20, in=160] (3.8,0);

	\end{scope}

	\begin{scope}[node distance=1.25cm,auto,>=latex', xshift=7.75cm]
		\tikzstyle{block}=[draw, fill=blue!20, minimum width=8mm,%
		minimum height=1cm]
		\node[yshift=1.25cm] (){\textbf{device training} (heterogeneous)};
		\node[yshift=-1.0cm, xshift=-3.1cm, text width=6cm, anchor=north west] (){\small layers [$1-3$] remain in their pretrained state and are not optimized by \ac{FL}};

		\node[draw, thick, fill opacity=0.1, minimum width=6cm, minimum height=2cm, fill=gray](){};

		\node[draw, thick, fill opacity=0.15, minimum width=2.75cm, minimum height=2cm, fill=green] at (1.625,0) {};

		\node[draw, minimum width=0.6cm, minimum height=0.5cm, fill=gray] at (-1.9,0.6) (b1) {};
		\node[draw,minimum width=0.6cm, minimum height=0.5cm, fill=gray, right=0.25cm of b1] (b2) {};
		\node[draw,minimum width=0.6cm, minimum height=0.5cm, fill=gray, right=0.25cm of b2] (b3) {};
		\node[draw,minimum width=0.6cm, minimum height=0.5cm, fill=red, right=0.25cm of b3] (b4) {};
		\node[draw,minimum width=0.6cm, minimum height=0.5cm, fill=red, right=0.25cm of b4] (b5) {};
		\node[draw,minimum width=0.6cm, minimum height=0.5cm, fill=red, right=0.25cm of b5] (b6) {};

		\draw[->] (b1) -- (b2);
		\draw[->] (b2) -- (b3);
		\draw[->] (b3) -- (b4);
		\draw[->] (b4) -- (b5);
		\draw[->] (b5) -- (b6);

		\node[draw,minimum width=0.6cm, minimum height=0.5cm, fill=gray, below=0.1cm of b1] (bb1) {};
		\node[draw,minimum width=0.6cm, minimum height=0.5cm, fill=gray, right=0.25cm of bb1] (bb2) {};
		\node[draw,minimum width=0.6cm, minimum height=0.5cm, fill=gray, right=0.25cm of bb2] (bb3) {};
		\node[draw,minimum width=0.6cm, minimum height=0.5cm, fill=gray, right=0.25cm of bb3] (bb4) {};
		\node[draw,minimum width=0.6cm, minimum height=0.5cm, fill=gray, right=0.25cm of bb4] (bb5) {};
		\node[draw,minimum width=0.6cm, minimum height=0.5cm, fill=red, right=0.25cm of bb5] (bb6) {};

		\draw[->] (bb1) -- (bb2);
		\draw[->] (bb2) -- (bb3);
		\draw[->] (bb3) -- (bb4);
		\draw[->] (bb4) -- (bb5);
		\draw[->] (bb5) -- (bb6);

		\node[draw, minimum width=0.6cm, minimum height=0.5cm, fill=gray, below=0.1cm of bb1] (bbb1) {};
		\node[draw, minimum width=0.6cm, minimum height=0.5cm, fill=gray, right=0.25cm of bbb1] (bbb2) {};
		\node[draw, minimum width=0.6cm, minimum height=0.5cm, fill=gray, right=0.25cm of bbb2] (bbb3) {};
		\node[draw, minimum width=0.6cm, minimum height=0.5cm, fill=gray, right=0.25cm of bbb3] (bbb4) {};
		\node[draw, minimum width=0.6cm, minimum height=0.5cm, fill=red, right=0.25cm of bbb4] (bbb5) {};
		\node[draw, minimum width=0.6cm, minimum height=0.5cm, fill=red, right=0.25cm of bbb5] (bbb6) {};
		
		\draw[->] (bbb1) -- (bbb2);
		\draw[->] (bbb2) -- (bbb3);
		\draw[->] (bbb3) -- (bbb4);
		\draw[->] (bbb4) -- (bbb5);
		\draw[->] (bbb5) -- (bbb6);

	\end{scope}

	\begin{scope}[node distance=1.25cm,auto,>=latex', xshift=14cm]
		\tikzstyle{block}=[draw, fill=blue!20, minimum width=8mm,%
		minimum height=1cm]

		\node[yshift=1.25cm] (){\textbf{aggregation}};
		\node[xshift=0.25cm, yshift=-1.0cm, text width=6cm, anchor= north] (){\small weighted aggregation (\cref{eq:aggregation1,eq:aggregation2})};

		\node[draw, thick, fill=gray, fill opacity=0.15, minimum width=3cm, minimum height=2cm, fill=green](){};
		\node[draw, minimum width=0.6cm, minimum height=0.5cm, fill=red] at (-0.85,0.5) (b1) {};
		\node[draw, minimum width=0.6cm, minimum height=1.0cm, fill=red] at (-0.05, 0.25) (b2) {};

		\node[minimum width=0.6cm, minimum height=1cm, fill=gray, fill opacity=0.5] at (-0.85, -0.25) (bb1) {};
		\node[minimum width=0.6cm, minimum height=0.5cm, fill=gray, fill opacity=0.5] at (-0.05, -0.5) (bb2) {};

		\node[draw, minimum width=0.6cm, minimum height=1.5cm, fill=red] at (+0.75,-0.0) (b3) {};

		\draw[->, ultra thick] (-1.8,0.25) to[out=140, in=40] (-2.9,0.25);
		\draw[<-, ultra thick] (-1.8,-0.25) to[out=-140, in=-40] (-2.9,-0.25);
	
	\end{scope}	
	
\end{tikzpicture}
    \caption{We propose \iac{FL} technique for cross-device \ac{FL} with heterogeneous devices that incorporates the availability of pretrained models. Unlike previous approaches, our technique retains some layers in their pretrained state.}
    \label{fig:overview}
\end{figure*}

\textbf{Heterogeneous freezing of pretrained \acp{NN}:} We consider each \ac{NN} architecture~$F_l$ to have~$l$ layers, where a specific number of these layers can be trained while others are frozen. We define~$F_l^t$ as an \ac{NN} architecture with~$t \leq l$ out of a total of~$l$ layers being trained. Specifically, in~$F_l^t$, we freeze the first~$l-t$ layers, while the remaining~$t$ layers with indices~$[l-t+1, \ldots, l]$ are being trained. Each architecture~$F_l$ can be trained with a different number of layers~$t \in [1,\ldots,l]$. We refer to~$l$ and~$t$ as a training \textit{configuration}. Each configuration has associated resource requirements. We label~$M_{F_l^t}$ as the memory required to train~$F_l^t$. Similarly, we label~$O_{F_l^t}$ as the computation operations (\acp{FLOP}) required, and~$U_{F_l^t}$ for communication upload, respectively. In general, a device~$c$ can only apply training to a given \ac{NN}-configuration~$F_l^t$ if
\begin{align}
    M_{F_l^t} \leq M_c \quad \land \quad U_{F_l^t} \leq U_c  \quad \land \quad O_{F_l^t} \leq O_c.
\end{align}

\subsection{\ac{NN}-architecture selection with heterogeneous devices}
We assume that each \ac{FL} device~$c$ should be capable in participating in the training. Consequently a feasible \ac{NN} structure~$F_l$ is required, where a device-specific~$t$ exits that allows all devices to apply training. Since there might be multiple architectures that allow this, we reduce the set~$\mathcal F$ to $\mathcal{F}_{\text{feasible}}$ by using
\begin{multline}
\label{eq:feasible}
\mathcal{F}_{\text{feasible}} \gets \{ F_l \in \mathcal{F} \ | \ (\exists t) [t \in [1,\ldots,l] \land \\ \forall c \in \mathcal C \ [M_{F_l^t} \leq M_c  \land U_{F_l^t} \leq U_c  \land O_{F_l^t} \leq O_c ]]\}
\end{multline}
Based on our observations from preliminary experiments (such as those presented in \cref{fig:motivation}), we propose the following \ac{NN} selection scheme. Specifically, we pick~$\hat{F_l}$ such that the average number of trained layers~$t_c$ by the devices~$c \in \mathcal{C}$ is maximized
\begin{align}
\label{eq:min2}
\textstyle
    \hat{F_l} = \text{argmax}_{F_l} \Big\{\frac{1}{\mathcal{|C|}} \sum_{c \in C} t_c \ | \ \forall F_l \in \mathcal{F}_{\text{feasible}}\Big\}.
\end{align} 
If there are two \acp{NN} that allow for the same average trained layers, we pick the NN with the most total layers $l$. This has specific implications depending on the constraint scheme. If a setting is purely peak memory or upload constrained, we observe that the associated cost of training $t$ layers is mostly independent of the total layers $l$ (as no activations have to be stored for frozen layers and no upload of frozen weights is required); hence, the \ac{NN} with the most layers is picked.
However, the number of computations (\acp{FLOP}) is impacted by~$t$ as well as by~$l$. This is because the calculation of the frozen forward pass contributes to the computation footprint (i.e., the cost of training~$t$ layers of a 6-layered \ac{NN} is lower than training~$t$ layers of a 9-layered \ac{NN}). We observe (\cref{fig:motivation}) that, generally, training more layers of a more shallow \ac{NN} maximizes the accuracy under a pure computation constraint. However, our scheme (\cref{eq:feasible,eq:min2}) also allow for a mixture of different constraints. We provide an ablation study for the heterogeneous case in \cref{subsec:ablation_study} to show that this strategy generally maximizes accuracy in computationally constrained cases.

We label the weights of~$F_l$ as~$\mathcal{W} = \{W_i \ | \ i \in [1, \ldots, l]\}$, where the weights of an individual layer are referred to as~$W_i$. Based on its constraints, a device~$c$ selects the number of trained layers~$t_c$ and applies training to~$\mathcal W^{(c)} = \{W_i \ | \ i \in [l-t_c+1, \ldots, l]\}$. Since layers~$[1,\ldots,l-t_c]$ remain frozen, their respective parameters do not have to be uploaded to the server, saving on communication costs. On the server, we apply layer-wise weighted averaging (as done in \cite{pfeiffer2023cocofl}) such that for each layer~$i$, the averaged weights~$W_i^{(r+1)}$ are determined by
\begin{align}
\textstyle
\label{eq:aggregation1}
\tilde{\mathcal W}_i = \bigcup_{c \in \mathcal C} \{W_i^{(r,c)} \ | \ \exists W_i^{(r,c)} \in \mathcal{W}^{(r,c)} \} \\
\textstyle
\label{eq:aggregation2}
W_i^{(r+1)} = \frac{1}{|\tilde{\mathcal W}_i|} \sum_{W_j \in \tilde{\mathcal W}_i} W_j + (1-\frac{1}{|\tilde{\mathcal W}_i|})\cdot W_i^{(r)},
\end{align} where $\tilde{\mathcal W}_i$ represents the set of weights of layer $i$, that are trained by the devices, and $W_i^{(r)}$ refers to the weights of layer $i$ from the previous round. The \ac{NN} selection and aggregation scheme is described in \cref{alg:algorithm}, and we provide an overview of individual components in \cref{fig:overview}.

\section{Experimental Evaluation}
\subsection{Hyperparameters and setting}
\label{subsec:hyperparameters}
\textbf{Pretraining:} We evaluate our technique and state of the art based on a set of pretrained models~$\mathcal{F}$. Specifically, we pretrain Transformers with layers~$l \in [3,6,9,12]$. Each \ac{NN} model~$F_l$ only varies in the number of layers. We use an embedding dimension of~$96$ and~$3$ heads per attention block for language modelling and $192$ and $6$ for \acp{VIT}. For language tasks, we use a sentencepiece tokenizer with a vocab size of~$8192$ and pretrain for~$750\text{K}$ mini-batch steps on OpenWebText~\cite{Gokaslan2019OpenWeb}, using a context length of~$256$ and batch size~$128$. For vision tasks, we use a \ac{VIT}~\cite{dosovitskiy2020image} with patch size of $4$ and a context length of $272$. We pretrain on a downscaled ($3\times64\times64$) version of ImageNet~\cite{deng2009imagenet} for~$500\text{K}$ steps with a batch size of~$256$. We apply data augmentation techniques such as random flipping, rotation, as well as random cropping. For both, vision and text domain, we use an initial learning rate of~$\eta=5\cdot10^{-4}$ and use a sinusoidal decay to~$\eta = 5\cdot10^{-5}$ ($1\cdot10^{-4}$ and $1\cdot10^{-5}$ for \ac{VIT}). We use AdamW~\cite{loshchilov2017decoupled} as optimizer ($\beta_1 = 0.9, \beta_2 = 0.95$), with weight decay of~$0.1$ for linear layers. In all cases, dropout of~$0.05$ is used.

\textbf{Federated training:}
We train a downstream task in a federated manner, where we distribute equal shares of Shakespeare from the Leaf benchmark~\cite{caldas1812leaf} (next token prediction), and CIFAR10, and CIFAR100~\cite{krizhevsky2009learning} to devices~$c \in \mathcal C$.  Consequently, each device~$c$ has a local private dataset $\mathcal{D}_c$. For CIFAR and Shakespeare, we use a total of~$|\mathcal{C}| = 100$ and~$10$ devices per round ($|\mathcal{C}^{(r)}| = 10$). We train for a total number of~$R=75$ rounds. In the case of Shakespeare, each device randomly picks a sentence using a context length of~$256$ from its local text, and trains for~$8$ batches with batch size~$32$. In the case of CIFAR, each device iterates once over its data samples using batch size~$32$. To have the same input resolution as the data used for pretraining, we upscale CIFAR data from \(3 \times 32 \times 32\) to \(3 \times 64 \times 64\). We distribute CIFAR100 and CIFAR10 in an non-\ac{IID} fashion with Dirichlet $\alpha=0.1$ and $\alpha=1.0$, respectively. We use the same optimizer setup as used in pretraining and apply learning rate decay from~$\eta = 1 \cdot 10^{-4}$ to~$\eta = 1 \cdot 10^{-5}$ (to $\eta = 1 \cdot 10^{-6}$ in case of Shakespeare). In case of Shakespeare, we use weight decay of $0.1$. For all federated training, dropout is set to~$0$. For each experiment, we report the average accuracy and standard deviation of 3 independent seeds after $R$ total rounds of training. For \ac{OURS}, we pick the \ac{NN} and number of layers to train based on \cref{eq:feasible}, and apply \cref{eq:aggregation1,eq:aggregation2} for aggregation. In all cases, the Transformer's embedding layers remain frozen, while the output layers (LayerNorm and linear layer) always receive training.

In total, we run approximately $1788$ federated experiments using PyTorch 2.1~\cite{PyTorch} using an internal cluster of NVIDIA A6000 GPUs with $\SI{50}{\giga\byte}$ of memory each (we train $\SI{894}{\hour}$ in total, with each experiment taking $\sim\SI{30}{\minute}$).

\textbf{Resource constraints}: We consider peak memory, communication, and computation as main constraints. Details about measuring a configuration's memory, communication, and computation footprint are given details in \cref{appdx:technical_details}.

\subsection{Ablation study}
To evaluate how the \ac{NN} selection~\cref{eq:min2} behaves when having heterogeneous devices, we mix two groups with computation constraints $[55,75]$ and $[75,100]$ GFLOPs with different rates. I.e., a rate of~$10\%$ refers to~$10\%$ of devices having a constraint of~$75$ while~$90\%$ have~$55$. For ratios between~$0\%$ and~$100\%$, we evaluate the accuracy on Shakespeare with all feasible Transformer architectures and what~$\hat F_l$ is picked by our technique based on the ratio.

\begin{figure}
    \centering
    \includegraphics[page=1]{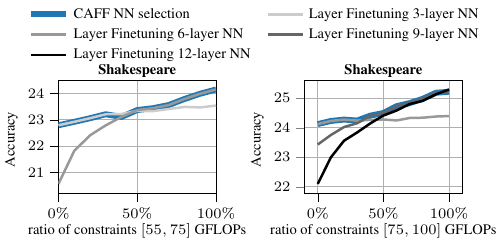}
    \caption{Ablation study of \ac{NN} selection (\cref{eq:min2}). We observe that picking the \ac{NN} that maximizes the average of trained layers maximizes the accuracy (blue).}
    \label{fig:ablation_study}
\end{figure}

We can observe in \cref{fig:ablation_study} that maximizing the average number of trained layers (as in \cref{eq:min2}) robustly picks the \ac{NN} that maximizes the accuracy. Depending on the average trained layers $t$ (which depends on the mixing ratio), \cref{eq:min2} switches from a 3 to a 6 and 6 to a 9-layered \ac{NN}, maximizing the accuracy.

\label{subsec:ablation_study}

\subsection{State of the art comparison}
\label{subsec:sota_comparions}
We compare \acs{OURS} against several state of the art techniques: Heterogeneous LoRA~\cite{cho2023heterogeneous}, FedHM~\cite{yao2021fedhm}, HeteroFL~\cite{diao2020heterofl}, FjORD~\cite{horvath2021fjord}, FedRolex~\cite{alam2022fedrolex}, and DepthFL~\cite{kim2023depthfl}. We provide details about the configuration of the state of the art in \cref{appdx:state-of-the-art}.

\subsection{Experimental results}
\label{subsec:experimental_results}

\begin{table*}[ht]
    \begingroup
    \caption{Accuracy for Shakespeare and CIFAR100 with homogeneous memory, upload, and computation constraints. The results show that, especially with memory constraints, \acs{OURS} (ours) outperforms the current state of the art. Only, in case of upload constraints, Heterog. LoRA shows higher effectiveness when utilizing available resources. We report $-$ if no $F_l$ is available.}
    \centering
\footnotesize
\setlength{\tabcolsep}{7.0pt}
\renewcommand{\arraystretch}{0.7}
    \begin{tabular}{l c c c c c c c c c}
       \toprule
        Setting &  \multicolumn{3}{c}{peak memory ($\si{\mega\byte}$)} & \multicolumn{3}{c}{upload ($\si{\mega\byte}$)} & \multicolumn{3}{c}{computation (GFLOPs)} \\\cmidrule(l{1pt}r{5pt}){2-4} \cmidrule(l{1pt}r{5pt}){5-7} \cmidrule(l{1pt}r{5pt}){8-10}
        \textbf{Shakespeare} &  500 & 700 & 900 & 4 & 6 & 8 & 60 & 80 & 100\\
         \midrule
         \acs{OURS} (ours) & \textbf{23.4}$\pm$\textbf{0.1} & \textbf{24.9}$\pm$\textbf{0.3} & \textbf{25.6}$\pm$\textbf{0.3} & 22.2$\pm$0.2 & \textbf{24.8}$\pm$\textbf{0.2} & \textbf{25.4}$\pm$\textbf{0.3} & \textbf{23.1$\pm$0.1} & \textbf{24.6$\pm$0.3} & \textbf{25.0$\pm$0.2}\\
         Heterog. LoRA  & 22.0$\pm$0.3 & 23.2$\pm$0.5 & 23.6$\pm$0.3 & \textbf{23.2}$\pm$\textbf{0.2} & 24.5$\pm$0.1 & 24.5$\pm$0.1& 22.0$\pm$0.3 & 23.2$\pm$0.5 & 23.6$\pm$0.3\\
         FedHM  & 18.5$\pm$0.3 & 19.0$\pm$0.2 & 19.0$\pm$0.2 & $-$ & $-$ & 19.0$\pm$0.2 & 19.0$\pm$0.2 & 19.0$\pm$0.2 & 19.0$\pm$0.2\\
         FedRolex & \phantom{0}1.3$\pm$0.0 & \phantom{0}4.5$\pm$0.3 & \phantom{0}4.5$\pm$0.3 & \phantom{0}3.1$\pm$0.1 & \phantom{0}4.5$\pm$0.3 & \phantom{0}4.5$\pm$0.3 & \phantom{0}4.5$\pm$0.3 & \phantom{0}4.5$\pm$0.3 & \phantom{0}4.5$\pm$0.3\\
         \midrule
         \textbf{CIFAR100} &  400 & 800 & 1200 & 8 & 12 & 16 & 100 & 150 & 200\\
         \midrule
         \acs{OURS} (ours)  & \textbf{56.6}$\pm$\textbf{0.2} &  \textbf{62.2}$\pm$\textbf{0.5} & \textbf{63.9}$\pm$\textbf{0.6} & 58.4$\pm$0.2 & 60.9$\pm$0.1 & \textbf{63.7}$\pm$\textbf{0.4} & \textbf{43.2}$\pm$\textbf{1.3} & \textbf{56.1}$\pm$\textbf{1.2} & \textbf{60.8}$\pm$\textbf{1.0}\\
         Heterog. LoRA  & 38.7$\pm$0.4 &  53.5$\pm$0.9 & 57.5$\pm$0.4 & \textbf{62.3}$\pm$\textbf{0.1} & \textbf{62.3}$\pm$\textbf{0.1} & 62.3$\pm$0.1 & 38.7$\pm$0.4 & 53.5$\pm$0.9 & 46.1$\pm$0.8 \\
         FedHM  & 22.7$\pm$0.4 & 32.0$\pm$0.6 & 37.3$\pm$1.3 & 41.0$\pm$0.2 & 41.0$\pm$0.2 & 41.0$\pm$0.2 & 37.3$\pm$1.3 & 41.0$\pm$0.2 & 41.0$\pm$0.2 \\
        FedRolex & \phantom{0}1.2$\pm$0.1 & \phantom{0}1.2$\pm$0.1 & \phantom{0}1.1$\pm$0.1 & \phantom{0}1.1$\pm$0.1 & \phantom{0}1.1$\pm$0.1 & \phantom{0}1.1$\pm$0.1 & \phantom{0}1.1$\pm$0.1 & \phantom{0}1.1$\pm$0.1 & \phantom{0}1.1$\pm$0.1\\      
         \bottomrule
    \end{tabular}
    \label{tab:results}
    \endgroup
\end{table*}

\begin{figure*}
    \centering
    \includegraphics[page=1]{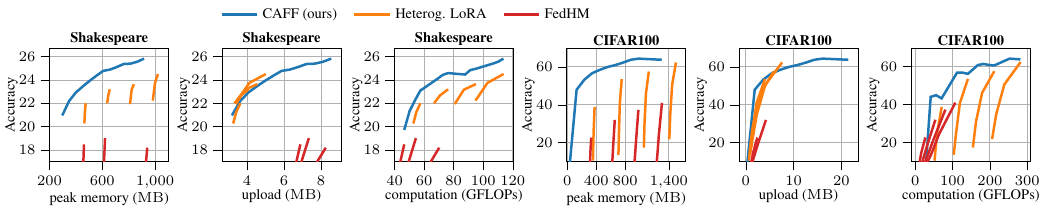}
    \caption{Visualization of homogeneous results for Shakespeare and CIFAR100. For \acs{OURS} (ours), we apply \ac{NN} selection based on \cref{eq:feasible,eq:min2}. For Heterog. LoRA and FedHM, we evaluate all \ac{NN} architectures and, for a given constraint, present the best performing in \cref{tab:results}.}
    \label{fig:homogeneous_results}
\end{figure*}

\textbf{Homogeneous results:} We evaluate our technique and Heterog. LoRA as well as FedHM and FedRolex in a homogeneous setting with Shakespeare and CIFAR100. Specifically, we run several experiments with homogeneous memory constraints of $500$, $700$, and $900$ ($400$, $600$, and $800$ for \ac{VIT}) \si{\mega\byte}, upload constraints of $4$, $6$, and $8$ ($8$, $12$, and $16$ for \ac{VIT}) \si{\mega\byte}, as well as computation constraints of $60$, $80$, and $100$ ($75$, $150$, $200$ for \ac{VIT}) GFLOPs. We omit results for HeteroFL, FjORD, and DepthFL, as in the homogeneous case they all degrade to vanilla FedAvg. For completeness, we visualize a range of constraints in \cref{fig:homogeneous_results}. To have a fair comparison, we evaluate the technique with pretrained transformers with $l \in [3,6,9,12]$ layers and present the respective highest gained accuracy in \cref{tab:results}. For our technique, we select the number of layers based on \cref{eq:min2}.

In general, we can observe that \acs{OURS} achieves higher accuracies in almost all considered constraint settings, particularly with memory constraints. Here it can be observed that, to reach a certain accuracy, LoRA requires $2-3\times$ more memory compared to \acs{OURS}. However, LoRA achieves higher accuracies when upload constraints are applied. Additionally, it can be observed that FedRolex, when training a larger pretrained model than the devices can fully train within a round~\cref{eq:fedrolex}, the accuracy heavily deteriorates. In \cref{fig:homogeneous_results}, we can see that with both Heterog. LoRA and FedHM lowering the rank for a specific \ac{NN} impacts upload and computation but have minimal effect on the required peak memory. In contrast, \acs{OURS} supports a wider range of applicability across all three constraints.

\textbf{Heterogeneous results:} To explore how our technique performs when devices have heterogeneous constraints, we explore a scenario where 50\% of devices have higher constraints than the other 50\% throughout all rounds. Specifically, we evaluate a setting an equal share of devices having memory constraints of $[400,600]$, $[600,800]$, and $[400,800]$ ($[200,400]$, $[400,600]$, and $[200,800]$ for \ac{VIT}) $\si{\mega\byte}$. Similarly, we evaluate $[2, 8], [4,8]$ ($[2,4]$ and $[2,8]$ $\si{\mega\byte}$ for \ac{VIT}) for upload constraints, and $[55, 75]$ and $[75,100]$ ($[75, 100]$ and $[125,200]$ for \ac{VIT}) GFLOPs for computation, respectively.

We observe that, across most evaluated scenarios, \acs{OURS} results in higher accuracy when devices have heterogeneous constraints. For \acs{OURS}, higher average resources lead to higher final accuracy, indicating that our technique efficiently uses available resources. However, with subset-training derived techniques, this is not always the case; for instance, with higher average computation in Shakespeare, many baselines actually perform worse or do not improve. For DepthFL we observe that the additional exit of the stronger devices significantly increases the upload and memory overhead, hence, for many evaluated scenarios (especially in case of language models), no suitable configuration is available.

\textbf{Can weaker devices make a contribution to the global model?} To evaluate if weaker devices can contribute to the model, we change the data distribution from non-\ac{IID} to a distribution that correlates with the devices' constraints (similar to \cite{pfeiffer2023cocofl}, using $\alpha=0.1$). Specifically, we distribute the data samples in CIFAR100 such that some classes are primarily available in a specific group, hence, either within the 50\% of weaker or 50\% of stronger devices. To evaluate how the global model performs exclusively on the weaker devices' data, we calculate the class-specific F1-score, and weigh it with the occurrence probability of the weaker devices' data. The results are given in~\cref{tab:results_heterogeneous}.

We can observe that using \acs{OURS}, the specific classes of the weaker devices are represented in the global model. However, when the variance of resources increases, weaker devices are less well represented (i.e., for memory constraints $[200,400]$ performs better than $[200,800]$ as stronger devices have more impact on the model). Contrary to that, we observe that the shared aggregation mechanism of Heterog. LoRA, FedRolex, HeteroFL, and FjORD is ineffective when such a data distribution is present, as the F1-score is 20-30 points less than when using \acs{OURS}. However, it can be seen, that in the aggregation mechanism of FedHM outperforms the other baselines. The only exception to that observation is when no suitable configuration for the baselines exists (labeled with *, \textdagger, and~\textdaggerdbl~in~\cref{tab:results_heterogeneous}), and the respective algorithms behave like in the homogeneous case (FedRolex, HeteroFL, and FjORD degrade to FedAvg in this case). For DepthFL, we see similar deterioration as in HeteroFL or FjORD. Be believe this is caused by the fact that weaker devices only train the early layers instead of the last layers and cannot sufficiently influence the stronger devices' exit.

\begin{table*}[h]
    \begingroup
    \caption{Accuracy for Shakespeare, CIFAR10/100. To measure fairness, we evaluate the F1-score specific to the weaker devices' data distribution. We observe that \acs{OURS} (ours) results in higher accuracies and better resource utilization, whereas state-of-the-art methods may even see accuracy deteriorate with increased resource availability.}
    \footnotesize
    \centering
    \renewcommand{\arraystretch}{0.7}
    \setlength{\tabcolsep}{8.5pt}
    \begin{tabular}{l l l l l l l l l l}
       \toprule
        \textbf{Shakespeare} &  \multicolumn{3}{c}{peak memory (\si{MB})} & \multicolumn{2}{c}{upload (\si{MB})} & \multicolumn{2}{c}{computation (GFLOPs)} \\\cmidrule(l{1pt}r{5pt}){2-4} \cmidrule(l{1pt}r{5pt}){5-6} \cmidrule(l{1pt}r{5pt}){7-8}
        Constraint &  [400,600] & [600,800] & [400,800] & [2,8] & [4,8]  & [55,75] & [75,100] \\
         \midrule
         \acs{OURS} (ours) & \textbf{23.9}$\pm$\textbf{0.5} & \textbf{25.1}$\pm$\textbf{0.4} & \textbf{24.8}$\pm$\textbf{0.3} & \textbf{24.6}$\pm$\textbf{0.4} & \textbf{25.3}$\pm$\textbf{0.5} & \textbf{23.3}$\pm$\textbf{0.4} & \textbf{24.6}$\pm$\textbf{0.4}\\
         Heterogenous LoRA  &$-$& 21.9$\pm$0.3\textsuperscript{\textdaggerdbl} & $-$ & 24.4$\pm$0.3  & 24.7$\pm$0.3 & 21.9$\pm$0.2  & 23.3$\pm$0.2  \\
         FedHM  & $-$ & 18.8$\pm$0.0\textsuperscript{\textdagger} &$-$ &$-$ & $-$ & 18.9$\pm$0.4 &19.3$\pm$0.2 \\
         FedRolex & 16.2$\pm$0.2 &  15.7$\pm$0.1 & 16.2$\pm$0.2 & $-$ &  16.3$\pm$0.3 &  16.2$\pm$0.2 & 15.7$\pm$0.1 \\
         HeteroFL & 22.1$\pm$0.4 & 23.0$\pm$0.2 & 22.1$\pm$0.4 &$-$ & 22.0$\pm$0.5 & 22.1$\pm$0.4 & 23.0$\pm$0.2\\
         FjORD & 19.8$\pm$0.3 & 21.5$\pm$0.1 & 19.8$\pm$0.3 & $-$& 19.8$\pm$0.2 & 19.8$\pm$0.3 & 21.5$\pm$0.1\\
         DepthFL & $-$ & $-$ & $-$ & $-$ & $-$ & $-$ & 21.7$\pm$0.4 \\
         \midrule
        \textbf{CIFAR100} &  [200,400] & [400,600] & [200,800] & [2,4] & [2,8]  & [75,100] & [125,200] \\
         \midrule
         \acs{OURS} (ours)  &  \textbf{52.7}$\pm$\textbf{0.1} & \textbf{58.8}$\pm$\textbf{0.2} & \textbf{57.2}$\pm$\textbf{0.5} & \textbf{50.6}$\pm$\textbf{0.2} & 54.1$\pm$0.4 & \textbf{42.3}$\pm$\textbf{1.7} & \textbf{60.2}$\pm$\textbf{1.1}\\
         Heterogenous LoRA  & $-$ & 38.9$\pm$0.9\textsuperscript{\textdaggerdbl} & $-$ & 45.0$\pm$1.1 & \textbf{55.2}$\pm$\textbf{0.5} & 38.9$\pm$0.9\textsuperscript{\textdaggerdbl} & 49.2$\pm$0.9\\
         FedHM  & $-$  & 23.5$\pm$1.6\textsuperscript{\textdagger} & $-$ & $-$ & 18.0$\pm$0.6 & 40.1$\pm$0.6\textsuperscript{\textdagger} & 31.6$\pm$1.3\\
         
         FedRolex & 33.9$\pm$1.0 & 40.7$\pm$0.8\textsuperscript{*} & 33.9$\pm$1.0 & $-$ & 32.7$\pm$0.6 & 40.7$\pm$0.8 \textsuperscript{*} & 44.7$\pm$2.6\\
         
         HeteroFL & 33.3$\pm$1.0 & 41.0$\pm$0.9\textsuperscript{*} & 33.3$\pm$1.0 & $-$ &  33.4$\pm$0.5 & 41.0$\pm$0.9\textsuperscript{*} & 45.6$\pm$1.9\\
         FjORD & 26.3$\pm$0.3 & 40.1$\pm$0.9\textsuperscript{*} & 26.3$\pm$0.3 & $-$ & 24.2$\pm$0.5 & 40.1$\pm$0.9\textsuperscript{*} & 37.0$\pm$1.2\\
         DepthFL & 26.8$\pm$1.2 & 41.0$\pm$0.9\textsuperscript{*} & 26.8$\pm$1.2 & $-$ & $-$ & 41.0$\pm$0.9\textsuperscript{*} & 28.7$\pm$1.3\\
         \midrule
         \textbf{CIFAR100 (F1-score weak)} &  [200,400] & [400,600] & [200,800] & [2,4] & [2,8]  & [75,100] & [125,200] \\
         \midrule
         \acs{OURS} (ours)  &  \textbf{29.8}$\pm$\textbf{3.4}  & \textbf{49.7}$\pm$\textbf{2.3} & \textbf{25.9}$\pm$\textbf{4.3} & \textbf{36.9}$\pm$\textbf{2.8} & \textbf{27.4}$\pm$\textbf{3.3} & \textbf{46.2}$\pm$\textbf{1.1}  & \textbf{40.4}$\pm$\textbf{4.6}\\
         Heterogenous LoRA  & $-$ & 40.2$\pm$2.9\textsuperscript{\textdaggerdbl} & $-$ & 10.4$\pm$3.8 & 13.5$\pm$4.4 & 40.2$\pm$2.9\textsuperscript{\textdaggerdbl} & 11.8$\pm$4.0\\
         FedHM  &  $-$ & 24.6$\pm$3.6\textsuperscript{\textdagger}  & $-$ & $-$ & 16.1$\pm$4.8 & 43.5$\pm$1.4\textsuperscript{\textdagger}  & 28.9$\pm$7.0 \\
         
         FedRolex & \phantom{0}8.1$\pm$2.9 & 43.9$\pm$3.6\textsuperscript{*} & \phantom{0}8.1$\pm$2.9 & $-$& \phantom{0}7.6$\pm$2.6 & 43.9$\pm$3.6\textsuperscript{*}  & 11.0$\pm$4.4\\
         
         HeteroFL & \phantom{0}8.0$\pm$2.9 & 43.9$\pm$3.4\textsuperscript{*} & \phantom{0}8.0$\pm$2.9 &$-$ & \phantom{0}7.8$\pm$2.8 & 43.9$\pm$3.4\textsuperscript{*}  &11.0$\pm$4.6 \\
         FjORD & \phantom{0}5.4$\pm$1.8 & 39.4$\pm$4.9\textsuperscript{*} & \phantom{0}5.4$\pm$1.8 & $-$ & \phantom{0}4.8$\pm$1.5 & 39.4$\pm$4.9\textsuperscript{*} & \phantom{0}7.7$\pm$3.4\\
         DepthFL & \phantom{0}5.2$\pm$1.6 & 43.9$\pm$3.4\textsuperscript{*} & \phantom{0}5.2$\pm$1.6 & $-$ & $-$ & 43.9$\pm$3.4\textsuperscript{*} & \phantom{0}6.0$\pm$2.1\\
         \midrule
          \textbf{CIFAR10} &  [200,400] & [400,600] & [200,800] & [2,4] & [2,8]  & [75,100] & [125,200] \\
         \midrule
         \acs{OURS} (ours)  &  \textbf{86.9}$\pm$\textbf{0.1} &  \textbf{89.3}$\pm$\textbf{0.1} & \textbf{89.8}$\pm$\textbf{0.1} & 85.5$\pm$0.3 & 87.8$\pm$0.1 & \textbf{82.4}$\pm$\textbf{0.2} & \textbf{91.1}$\pm$\textbf{0.2}\\
         Heterogenous LoRA  & $-$ & 80.1$\pm$0.1\textsuperscript{\textdaggerdbl} & $-$ & \textbf{89.4}$\pm$\textbf{0.6} & \textbf{90.4}$\pm$\textbf{0.5} & 80.1$\pm$0.1\textsuperscript{\textdaggerdbl} & 87.3$\pm$0.5\\
         FedHM  &  $-$ & 71.3$\pm$1.3\textsuperscript{\textdagger} & $-$ & $-$ & 64.8$\pm$1.0 & 81.9$\pm$0.2\textsuperscript{\textdagger} & 78.5$\pm$0.6\\
         
         FedRolex & 76.8$\pm$1.1 & 80.5$\pm$1.2\textsuperscript{*} & 76.8$\pm$1.1 & $-$  &   77.2$\pm$1.0 & 80.5$\pm$1.2\textsuperscript{*} & 86.8$\pm$0.4 \\
        
         HeteroFL & 76.9$\pm$1.1 & 80.6$\pm$1.2\textsuperscript{*} & 76.9$\pm$1.1 & $-$ &77.5$\pm$1.1 & 80.6$\pm$1.2\textsuperscript{*} & 87.1$\pm$0.3\\
         FjORD & 74.0$\pm$1.3 & 80.5$\pm$0.2\textsuperscript{*} & 74.0$\pm$1.3 & $-$ &72.6$\pm$2.0 & 80.5$\pm$0.2\textsuperscript{*} & 83.9$\pm$0.0\\
         DepthFL & 74.3$\pm$1.1 & 80.5$\pm$0.2\textsuperscript{*} & 74.3$\pm$1.1 & $-$ & $-$ & 80.5$\pm$0.2\textsuperscript{*}  & 85.9$\pm$0.2\\
         \bottomrule
    \end{tabular}
    \label{tab:results_heterogeneous}
    \scriptsize
    \begin{tablenotes}
        \item * Degrades to vanilla FedAvg, \textdagger~degrades to homogeneous FedHM, \textdaggerdbl~degrades to vanilla LoRA, $-$ indicates no suitable $F_l$.
    \end{tablenotes}
    \normalsize
    \endgroup
\end{table*}

\emph{In summary, we find that \acs{OURS} consistently outperforms existing baselines when using pretrained tiny Transformers suitable for cross-device \ac{FL}. However, in upload-constrained cases, Heterog. LoRA can outperform \acs{OURS}. When devices have heterogeneous constraints, \acs{OURS} results in higher accuracies and better resource utilization, whereas state-of-the-art methods may even see accuracy deteriorate with increased resource availability. Additionally, \acs{OURS} ensures fairer device contributions to the global model, especially when some class information is only available on weaker devices, which struggle to contribute under state-of-the-art methods.}

\section{Conclusion}
\label{sec:conclusion}
In this work, we studied efficient training of pretrained tiny \acp{NN} in cross-device \ac{FL}. We discovered that existing techniques like LoRA do not provide a good trade-off between resource usage and accuracy when considering peak memory and computation. Given a set of pre-trained \ac{NN} architectures to choose from, we propose \ac{OURS}, an \ac{NN} selection scheme and layer finetuning approach that outperforms LoRA and other state-of-the-art techniques, improving accuracy and fairness in resource-constrained scenarios. We believe that our technique, particularly the \ac{NN} selection scheme, contributes to practical design considerations for building efficient \ac{FL} systems with device heterogeneity in mind. By leveraging pretrained models, we reduce the total resources required for training, thereby alleviating the training burden on \ac{FL} devices. Moreover, our \ac{NN} selection and aggregation scheme promotes higher fairness by ensuring that data from weaker devices is better represented.

\printbibliography[]

@InProceedings{diao2020heterofl,
  author        = {Diao, Enmao and Ding, Jie and Tarokh, Vahid},
  title         = {HeteroFL: Computation and communication efficient federated learning for heterogeneous clients},
  booktitle     = {International Conference on Learning Representations (ICLR)},
  year          = {2020},
}

@Article{caldas2018expanding,
  author        = {Caldas, Sebastian and Kone{\v{c}}ny, Jakub and McMahan, H Brendan and Talwalkar, Ameet},
  title         = {Expanding the reach of federated learning by reducing client resource requirements},
  journal       = {arXiv preprint arXiv:1812.07210},
  year          = {2018},
}

@Misc{yu2020heterogeneous,
  author        = {Fuxun Yu and Weishan Zhang and Zhuwei Qin and Zirui Xu and Di Wang and Chenchen Liu and Zhi Tian and Xiang Chen},
  title         = {Heterogeneous Federated Learning},
  year          = {2020},
  archiveprefix = {arXiv},
  eprint        = {2008.06767},
  primaryclass  = {cs.LG},
}

@Article{caldas1812leaf,
  author  = {Caldas, S and Wu, P and Li, T and Konecn{\`y}, J and McMahan, HB and Smith, V and Talwalkar, A},
  title   = {Leaf: A benchmark for federated settings. arXiv 2018},
  journal = {arXiv preprint arXiv:1812.01097},
  year    = {2019},
}

@Article{yuan2020federated,
  author        = {Yuan, Binhang and Ge, Song and Xing, Wenhui},
  title         = {A federated learning framework for healthcare IoT devices},
  journal       = {arXiv preprint arXiv:2005.05083},
  year          = {2020},
}

@Article{chen2020fedhealth,
  author        = {Chen, Yiqiang and Qin, Xin and Wang, Jindong and Yu, Chaohui and Gao, Wen},
  title         = {Fedhealth: A federated transfer learning framework for wearable healthcare},
  journal       = {IEEE Intelligent Systems},
  year          = {2020},
  volume        = {35},
  number        = {4},
  pages         = {83--93},
  publisher     = {IEEE},
}

@Article{posner2021federated,
  author        = {Posner, Jason and Tseng, Lewis and Aloqaily, Moayad and Jararweh, Yaser},
  title         = {Federated Learning in Vehicular Networks: Opportunities and Solutions},
  journal       = {IEEE Network},
  year          = {2021},
  publisher     = {IEEE},
}

@Article{ciftler2020federated,
  author        = {Ciftler, Bekir Sait and Albaseer, Abdullatif and Lasla, Noureddine and Abdallah, Mohamed},
  title         = {Federated learning for localization: A privacy-preserving crowdsourcing method},
  journal       = {arXiv preprint arXiv:2001.01911},
  year          = {2020},
}

@Misc{krizhevsky2009learning,
  author    = {Krizhevsky, Alex and Hinton, Geoffrey and others},
  title     = {Learning multiple layers of features from tiny images},
  year      = {2009},
  publisher = {Citeseer},
}

@Article{horvath2021fjord,
  author        = {Horvath, Samuel and Laskaridis, Stefanos and Almeida, Mario and Leontiadis, Ilias and Venieris, Stylianos I and Lane, Nicholas D},
  title         = {FjORD: Fair and Accurate Federated Learning under heterogeneous targets with Ordered Dropout},
  journal       = {arXiv preprint arXiv:2102.13451},
  year          = {2021},
}

@InCollection{PyTorch,
  author    = {Paszke, Adam and Gross, Sam and Massa, Francisco and Lerer, Adam and Bradbury, James and Chanan, Gregory and Killeen, Trevor and Lin, Zeming and Gimelshein, Natalia and Antiga, Luca and Desmaison, Alban and Kopf, Andreas and Yang, Edward and DeVito, Zachary and Raison, Martin and Tejani, Alykhan and Chilamkurthy, Sasank and Steiner, Benoit and Fang, Lu and Bai, Junjie and Chintala, Soumith},
  title     = {PyTorch: An Imperative Style, High-Performance Deep Learning Library},
  booktitle = {Advances in Neural Information Processing Systems 32},
  publisher = {Curran Associates, Inc.},
  year      = {2019},
  editor    = {H. Wallach and H. Larochelle and A. Beygelzimer and F. d\textquotesingle Alch\'{e}-Buc and E. Fox and R. Garnett},
  pages     = {8024--8035},
  url       = {http://papers.neurips.cc/paper/9015-pytorch-an-imperative-style-high-performance-deep-learning-library.pdf},
}

@Article{yao2021fedhm,
  author  = {Yao, Dezhong and Pan, Wanning and Wan, Yao and Jin, Hai and Sun, Lichao},
  title   = {FedHM: Efficient Federated Learning for Heterogeneous Models via Low-rank Factorization},
  journal = {arXiv preprint arXiv:2111.14655},
  year    = {2021},
}

@InProceedings{alam2022fedrolex,
  author    = {Alam, Samiul and Liu, Luyang and Yan, Ming and Zhang, Mi},
  title     = {FedRolex: Model-Heterogeneous Federated Learning with Rolling Sub-Model Extraction},
  booktitle = {Advances in Neural Information Processing Systems},
  year      = {2022},
}

@InProceedings{mei2022resource,
  author    = {Mei, Yiqun and Guo, Pengfei and Zhou, Mo and Patel, Vishal},
  title     = {Resource-Adaptive Federated Learning with All-In-One Neural Composition},
  booktitle = {Advances in Neural Information Processing Systems},
  year      = {2022},
}

@inproceedings{
cho2023heterogeneous,
title={Heterogeneous Lo{RA} for Federated Fine-tuning of On-device Foundation Models},
author={Yae Jee Cho and Luyang Liu and Zheng Xu and Aldi Fahrezi and Matt Barnes and Gauri Joshi},
booktitle={International Workshop on Federated Learning in the Age of Foundation Models in Conjunction with NeurIPS 2023},
year={2023},
url={https://openreview.net/forum?id=EmV9sGpZ7q}
}

@inproceedings{
babakniya2023slora,
title={{SL}o{RA}: Federated Parameter Efficient Fine-Tuning of Language Models},
author={Sara Babakniya and Ahmed Elkordy and Yahya Ezzeldin and Qingfeng Liu and Kee-Bong Song and MOSTAFA EL-Khamy and Salman Avestimehr},
booktitle={International Workshop on Federated Learning in the Age of Foundation Models in Conjunction with NeurIPS 2023},
year={2023},
url={https://openreview.net/forum?id=06quMTmtRV}
}

@article{hu2021lora,
  title={Lora: Low-rank adaptation of large language models},
  author={Hu, Edward J and Shen, Yelong and Wallis, Phillip and Allen-Zhu, Zeyuan and Li, Yuanzhi and Wang, Shean and Wang, Lu and Chen, Weizhu},
  journal={arXiv preprint arXiv:2106.09685},
  year={2021}
}

@misc{houlsby2019adapter,
      title={Parameter-Efficient Transfer Learning for NLP}, 
      author={Neil Houlsby and Andrei Giurgiu and Stanislaw Jastrzebski and Bruna Morrone and Quentin de Laroussilhe and Andrea Gesmundo and Mona Attariyan and Sylvain Gelly},
      year={2019},
      eprint={1902.00751},
      archivePrefix={arXiv},
      primaryClass={cs.LG}
}

@article{radford2019language,
  title={Language Models are Unsupervised Multitask Learners},
  author={Radford, Alec and Wu, Jeff and Child, Rewon and Luan, David and Amodei, Dario and Sutskever, Ilya},
  year={2019}
}

@article{pfeiffer2023federated,
author = {Pfeiffer, Kilian and Rapp, Martin and Khalili, Ramin and Henkel, J\"{o}rg},
title = {Federated Learning for Computationally Constrained Heterogeneous Devices: A Survey},
year = {2023},
issue_date = {December 2023},
publisher = {Association for Computing Machinery},
address = {New York, NY, USA},
volume = {55},
number = {14s},
issn = {0360-0300},
url = {https://doi.org/10.1145/3596907},
doi = {10.1145/3596907},
journal = {ACM Comput. Surv.},
month = {jul},
articleno = {334},
numpages = {27}
}

@article{
pfeiffer2023cocofl,
title={CoCo{FL}: Communication- and Computation-Aware Federated Learning via Partial {NN} Freezing and Quantization},
author={Kilian Pfeiffer and Martin Rapp and Ramin Khalili and Joerg Henkel},
journal={Transactions on Machine Learning Research},
issn={2835-8856},
year={2023},
url={https://openreview.net/forum?id=XJIg4kQbkv},
note={}
}

@inproceedings{wang2022progfed,
  title={ProgFed: effective, communication, and computation efficient federated learning by progressive training},
  author={Wang, Hui-Po and Stich, Sebastian and He, Yang and Fritz, Mario},
  booktitle={International Conference on Machine Learning},
  pages={23034--23054},
  year={2022},
  organization={PMLR}
}

@inproceedings{
pfeiffer2023aggregating,
title={Aggregating Capacity in {FL} through Successive Layer Training for Computationally-Constrained Devices},
author={Kilian Pfeiffer and Ramin Khalili and Joerg Henkel},
booktitle={Thirty-seventh Conference on Neural Information Processing Systems},
year={2023},
url={https://openreview.net/forum?id=nXNsqB4Yr1}
}

@inproceedings{
setayesh2023perfedmask,
title={PerFedMask: Personalized Federated Learning with Optimized Masking Vectors},
author={Mehdi Setayesh and Xiaoxiao Li and Vincent W.S. Wong},
booktitle={The Eleventh International Conference on Learning Representations },
year={2023},
url={https://openreview.net/forum?id=hxEIgUXLFF}
}

@article{malan2023automatic,
  title={Automatic Layer Freezing for Communication Efficiency in Cross-Device Federated Learning},
  author={Malan, Erich and Peluso, Valentino and Calimera, Andrea and Macii, Enrico and Montuschi, Paolo},
  journal={IEEE Internet of Things Journal},
  year={2023},
  publisher={IEEE}
}

@misc{Gokaslan2019OpenWeb,  
	title={OpenWebText Corpus},
	author={Aaron Gokaslan and Vanya Cohen},
	howpublished={\url{http://Skylion007.github.io/OpenWebTextCorpus}}, 
	year={2019}
}

@article{dosovitskiy2020image,
  title={An image is worth 16x16 words: Transformers for image recognition at scale},
  author={Dosovitskiy, Alexey and Beyer, Lucas and Kolesnikov, Alexander and Weissenborn, Dirk and Zhai, Xiaohua and Unterthiner, Thomas and Dehghani, Mostafa and Minderer, Matthias and Heigold, Georg and Gelly, Sylvain and others},
  journal={arXiv preprint arXiv:2010.11929},
  year={2020}
}

@inproceedings{deng2009imagenet,
  title={Imagenet: A large-scale hierarchical image database},
  author={Deng, Jia and Dong, Wei and Socher, Richard and Li, Li-Jia and Li, Kai and Fei-Fei, Li},
  booktitle={2009 IEEE conference on computer vision and pattern recognition},
  pages={248--255},
  year={2009},
  organization={Ieee}
}

@article{loshchilov2017decoupled,
  title={Decoupled weight decay regularization},
  author={Loshchilov, Ilya and Hutter, Frank},
  journal={arXiv preprint arXiv:1711.05101},
  year={2017}
}

@inproceedings{li2020train,
  title={Train big, then compress: Rethinking model size for efficient training and inference of transformers},
  author={Li, Zhuohan and Wallace, Eric and Shen, Sheng and Lin, Kevin and Keutzer, Kurt and Klein, Dan and Gonzalez, Joey},
  booktitle={International Conference on machine learning},
  pages={5958--5968},
  year={2020},
  organization={PMLR}
}

@article{touvron2023llama,
  title={Llama: Open and efficient foundation language models},
  author={Touvron, Hugo and Lavril, Thibaut and Izacard, Gautier and Martinet, Xavier and Lachaux, Marie-Anne and Lacroix, Timoth{\'e}e and Rozi{\`e}re, Baptiste and Goyal, Naman and Hambro, Eric and Azhar, Faisal and others},
  journal={arXiv preprint arXiv:2302.13971},
  year={2023}
}

@inproceedings{
kim2023depthfl,
title={Depth{FL} : Depthwise Federated Learning for Heterogeneous Clients},
author={Minjae Kim and Sangyoon Yu and Suhyun Kim and Soo-Mook Moon},
booktitle={The Eleventh International Conference on Learning Representations },
year={2023},
url={https://openreview.net/forum?id=pf8RIZTMU58}
}

@inproceedings{hu2021unit,
  title={Unit: Multimodal multitask learning with a unified transformer},
  author={Hu, Ronghang and Singh, Amanpreet},
  booktitle={Proceedings of the IEEE/CVF international conference on computer vision},
  pages={1439--1449},
  year={2021}
}

@article{wang2022git,
  author = {Wang, Jianfeng and Yang, Zhengyuan and Hu, Xiaowei and Li, Linjie and Lin, Kevin and Gan, Zhe and Liu, Zicheng and Liu, Ce and Wang, Lijuan},
  journal = {Trans. Mach. Learn. Res.},
  title = {GIT: A Generative Image-to-text Transformer for Vision and Language.},
  url = {http://dblp.uni-trier.de/db/journals/tmlr/tmlr2022.html#WangYHLLGLLW22},
  volume = 2022,
  year = 2022
}

\appendix
\section{Technical details of constraint evaluation}
\label{appdx:technical_details}
\textbf{Peak Memory:} Peak memory is measured by summing the three major components that impact peak memory: the number of weights that must be kept in memory, the optimizer state of AdamW, and the activations that must be stored in memory. Activations that must be kept in memory are collected by traversing the compute graph during backpropagation in PyTorch~\cite{PyTorch}.

\textbf{Communication:} The amount of data (in \si{\mega\byte}) that needs to be uploaded by a device is computed by tracking the parameters that change during training on the device (i.e., unfrozen parameters). Unchanged parameters do not need to be uploaded.

\textbf{Computation:} The number of GFLOPs required for a single mini-batch during training is calculated by determining the forward and backward operations needed for major operations such as linear layers, dot products, addition, and layer normalization,

\subsection{Training memory components}
\begin{figure}
    \centering
    \includegraphics[page=1]{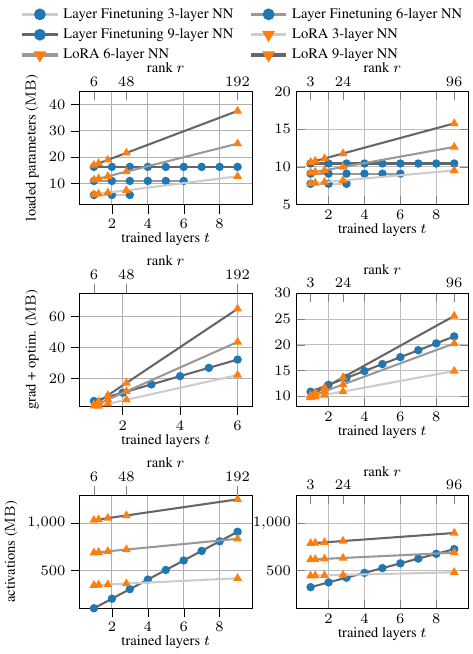}
    \caption{Memory components of layer fintuning and LoRA for language models (right) and \acp{VIT} (left) with 3, 6, and 9 layers.}
    \label{fig:memory-components}
\end{figure}
We analyze the different memory components involved in training with layer freezing as used in \acs{OURS} and baselines. The components \textit{parameters}, \textit{gradients and optimizer states}, and \textit{activations} are visualized for layer freezing and LoRA in \cref{fig:memory-components}. It can be observed that LoRA consistently requires more memory for loaded parameters compared to layer finetuning, due to the additional low-rank adapters. Conversely, LoRA demands less memory for gradients and optimizer states. However, activation memory constitutes the majority of the overall memory footprint for the evaluated transformer sizes. In this context, LoRA does not reduce activation memory since activations must be stored in memory for backpropagation, even when the main parameters of the transformer are frozen.

In contrast, with layer finetuning, the memory required for activations, gradients, and optimizer states is independent of the total number of layers and decreases linearly with the number of frozen layers. LoRA is very effective when gradients dominate the memory footprint. For large models (i.e., those with a large embedding dimension), this is the case, as gradients grow quadratically with the embedding dimension, while the activations only linearly depend on the embedding dimension.

\section{State of the art comparison}
\label{appdx:state-of-the-art}

\textbf{Heterogeneous LoRA~\cite{cho2023heterogeneous}:} LoRA applies a low-rank adapter to linear operations in the \ac{NN}'s layers. To support heterogeneity, the low-rank adapters are aggregated similarly to HeteroFL~\cite{diao2020heterofl}. We freeze the token and position embedding table similar to layer finetuning. All LayerNorm modules receive training. We fully train the output linear layer as we have seed bad performance when using low-rank adapters.

\textbf{FedHM~\cite{yao2021fedhm}:} In FedHM, a lower complexity model is created by applying a \ac{SVD} to linear layers~$w \in \mathbb R^{P\times Q}$ such that~$U, S, V^T = \text{svd}(w)$.
Using a lower rank~$z$, an approximation of~$w \sim U\cdot\text{diag}(S)\cdot V^T$ can be constructed using two consecutive linear layers with~$U\cdot\text{diag}(S) \in \mathbb R^{P\times z}$ and~$V^T \in \mathbb R^{z \times Q}$.
We pick the \ac{NN} structure so that the least constrained device can fully train the \ac{NN} without using \ac{SVD}. All devices with less resources pick~$z$ s.t.
\begin{equation}
    z_s = \text{max}(z) \quad \text{s.t.} \quad M_{F^z} \leq M_c  \land U_{F^z} \leq U_c  \land O_{F^z} \leq O_c.
\end{equation}
The parameters $w$ are reconstructed on the server for aggregation by using $U\cdot\text{diag}(S)\cdot V^T$.

\textbf{HeteroFL:~\cite{diao2020heterofl}:} HeteroFL is a state-of-the-art technique that enables heterogeneous devices to train a common \ac{NN} model. This is achieved by training a subset of the full \ac{NN} architecture. Let~$w \in \mathbb R^{P \times Q}$ represent a linear layer in a Transformer \ac{NN}. To lower the resources, constrained devices scale down the \ac{NN} by using a subset of~$w$ using a scale factor~$s \in (0, 1]$, such that~$\tilde{w} \in \mathbb R ^{\lfloor sP \rfloor \times \lfloor sQ \rfloor}$. Scaling down linear layers with~$s$ results in a quadratic reduction in parameters (hence upload and computation), but only in a linear reduction in memory (as activations make up for most required memory, which decreases linearly with~$s$. In HeteroFL, constrained devices receive always the same fixed subset of the full weights~$w$. We introduce~$I$ as the set of indices, that are used to create a subset. In HeteroFL, a device $c$ with a constraint~$s$ uses a subset of size~$\lfloor sP \rfloor \times \lfloor sQ \rfloor$ by using~$I_Q \in \{i | 0 \leq i < \lfloor sQ \rfloor -1\}$ as output indices and~$I_P \in \{i | 0 \leq i < \lfloor sP \rfloor -1 \}$ as input indices. HeteroFL requires that a share of \ac{FL} devices are capable of training the full \ac{NN} weights. Therefore, we select the \ac{NN} structure with the most layers that can still be fully trained ($s=1$) by a single participating device.
Remaining devices create a submodel $\hat F^s$ that supports their constraint by using 
\begin{align}
    s_c = \text{max}(s) \quad \text{s.t.} \quad M_{\hat F^s} \leq M_c  \land U_{\hat F^s} \leq U_c  \land O_{\hat F^s} \leq O_c.
\end{align}

\textbf{FjORD~\cite{horvath2021fjord}:} FjORD applies the same strategy as HeteroFL with respect to indices. However, a constrained device switches on a mini-batch level between different levels of~$s$.

\textbf{FedRolex~\cite{alam2022fedrolex}:} Here, devices use a subset similarly to HeteroFL. However, while in HeteroFL, a constrained device always trains the same slice of weights~$w$, in FedRolex, a rolling window approach is used. Output indices are selected per round~$r$, such that for a linear layer~$w$, $I$ is selected using
\begin{align}
\label{eq:fedrolex}
    I^{(r)} =
    \{ \hat r, \hat r+1, \ldots, r + \lfloor s_c Q\rfloor -1 \} 
\end{align} in case $\hat r + \lfloor s_c Q \rfloor \leq Q$.
\begin{align}
    \{ \hat r, \hat r+1, \ldots, Q -1 \} \cup \{0,\ldots, \hat r + \lfloor s_c Q  \rfloor - 1 - Q \} 
\end{align}
otherwise, where $\hat r = r \ \text{mod} \ Q$. Thereby, FedRolex eventually trains all weights of an \ac{NN}, even if no device can fully train all parameters within a round.

\textbf{DepthFL~\cite{kim2023depthfl}:} As we evaluate devices within two distinct resource groups, we configure DepthFL such that the weaker devices use a single early exit, while the stronger devices use the same early exit as the weaker devices, as well as the last exit of the \ac{NN} structure. On the server, we evaluate the accuracy based on the last exit of the \ac{NN}.

\end{document}